\documentclass[conference]{IEEEtran}
\IEEEoverridecommandlockouts
\usepackage{cite}
\usepackage{amsmath,amssymb,amsfonts}
\usepackage{algorithmic}
\usepackage{graphicx}
\usepackage{textcomp}
\usepackage{xcolor}
\usepackage{adjustbox}
\usepackage{multirow}
\def\BibTeX{{\rm B\kern-.05em{\sc i\kern-.025em b}\kern-.08em
    T\kern-.1667em\lower.7ex\hbox{E}\kern-.125emX}}
\begin{document}

\title{Smart Feature is What You Need
}

\author{\IEEEauthorblockN{1\textsuperscript{st} Zhaoxin Hu}
\IEEEauthorblockA{\textit{Beijing University of Technology}\\
Beijing, China \\
hzx75002@emails.bjut.edu.cn}
\and
\IEEEauthorblockN{2\textsuperscript{nd} Keyan Ren}
\IEEEauthorblockA{\textit{Beijing University of Technology} \\
Beijing, China \\
keyanren@bjut.edu.cn}
}

\maketitle

\begin{abstract}
Lack of shape guidance and label jitter caused by information deficiency of weak label are the main problems in 3D weakly-supervised object detection. Current weakly-supervised models often use heuristics or assumptions methods to infer information from weak labels without taking advantage of the inherent clues of weakly-supervised and fully-supervised methods, thus it is difficult to explore a method that combines data utilization efficiency and model accuracy. In an attempt to address these issues, we propose a novel plug-and-in point cloud feature representation network called Multi-scale Mixed Attention (MMA). MMA utilizes adjacency attention within neighborhoods and disparity attention at different density scales to build a feature representation network. The smart feature representation obtained from MMA has shape tendency and object existence area inference, which can constrain the region of the detection boxes, thereby alleviating the problems caused by the information default of weak labels. Extensive experiments show that in indoor weak label scenarios, the fully-supervised network can perform close to that of the weakly-supervised network merely through the improvement of point feature by MMA. At the same time, MMA can turn waste into treasure, reversing the label jitter problem that originally interfered with weakly-supervised detection into the source of data enhancement, strengthening the performance of existing weak supervision detection methods. Our code is available at https://github.com/hzx-9894/MMA.

\end{abstract}


\section{Introduction}
3D object detection has been extensively studied in the object detection community due to the widespread use of commercial 3D scanners\cite{3D1,3D2,3D3}. While capturing point clouds from real-world scenes becomes convenient and affordable, the complex 3D scenarios and redundant points make 3D object detection a very challenging task. 

In order to develop reliable 3D detection capabilities, fully supervised approaches require training on enough labeled annotations. Compared with the 2D labels, the fine-grained 3D annotations need to define the pitch, yaw, and roll angles, which is extremely labor-intensive and time-consuming. Relevant research shows that the time cost of labeling a well-annotated 3D data is about 100 seconds, which is much higher than the 5 to 10 seconds of labeling a 2D label \cite{BR}. Expensive annotation costs hinder the expansion of 3D data scale, making 3D object detection still far from large-scale applications. With the advent of weakly-supervised 3D object detection \cite{BR,WS3D,weakm3d,VS3D}, low-cost weakly-labeled training data can be learned for the detection model, which alleviates a huge amount of computation and annotation expenses. Leveraging this know-how, weakly-supervised detection enables the application of 3D object detection in intricate, variable, and data-intensive real-world scenarios. 

However, the research on weakly-supervised 3D object detection is not completely smooth sailing. Two main problems plaguing the research on weakly-supervised detection. On the one hand, the current 3D weakly-supervised model is often a complete system with a series of interconnected designs. The existing methods usually use heuristics or assumptions to infer 3D detection from limited data, thus each part of the learning framework needs to be tightly coupled, making it hard to flexibly and conveniently embed it into other networks. "Weakly-supervised detector is designed only for weak supervision", there is currently a lack of a generic and sharing method.

On the other hand, weak labels inevitably involve a large amount of ambiguous and noisy information in each training instance. More specifically, weak labels are usually part of precise labels, such as the center of the object. There are two major issues with these labels. The first issue is that weak labels lack a lot of information compared to precise labels, especially the shape guidance brought by the 3D detection boxes. This makes the detection boxes obtained by weak supervision often have the wrong shape and size. The researchers have found that weak supervision often performs similarly to fully supervised performance under low IoU requirements, while the gap is large under high IoU requirements \cite{sceneweak}. The second issue is that due to the simplification of labels, the information on weak labels tends to be unified, which makes the diversification and mutually complementary nature of accurate labels missing. When labels jitter, detection performance will be greatly affected. Widely studied have been developed for this issue. A common approach is to design center refinement and anti-jitter modules to specifically deal with these problems. Obviously, designing a processing module can indeed solve the problem, but it will undoubtedly aggravate the coupling. Some works consider directly guiding the generation of 3D detection boxes through point-label alignment, and have achieved good results. In fact, if we can get a suitable feature representation from a point cloud that specifies the existence area of the object, we could constrain the generation of the bounding boxes, easing the problem of the lack of shape guidance and label jitter. Since the feature representation is the common stage of most methods, this is actually a low-coupling method that also effectively solves problem one.

At present, due to the gradual expansion of research in the visual field of Transformer, a classic model in natural language processing, some research has begun to focus on using Transformer to extract point cloud features. Networks such as PCT and PT \cite{PCT,PT} use Transformer to aggregate global or local point cloud information to assist feature extraction. However, for the field of weakly-supervised detection, researchers still prefer to use traditional PointNet++ \cite{pointnet2}. PointNet++ is indeed a powerful method, however, relying solely on PointNet++ cannot fully meet the needs of determining the existence area and general shape of objects.

\begin{figure*}[htbp]
  \centering
  \includegraphics[width=0.8\textwidth]{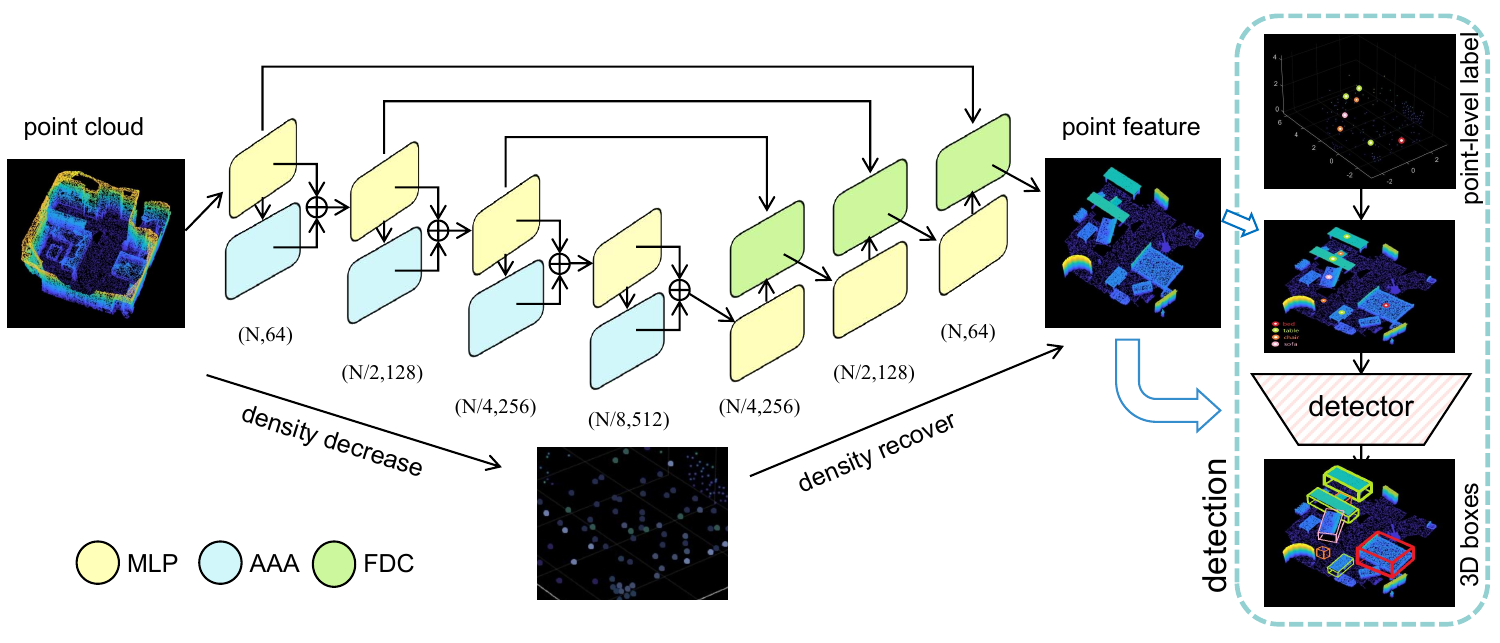}
  \caption{The overall structure of MMA. The left side is the MMA, and the right side within dotted line is the schematic diagram of weakly-supervised detection using MMA. MLP stands for multi-layer perceptron, AAA stands for Adjacency Attention Aggregation layer, and FDC stands for Feature Disparity Calculation layer.}
  \label{structure}
\end{figure*}

In this paper, we propose a novel point cloud feature representation network from off-the-shelf 3D detection models called Multi-scale Mixed Attention (MMA) to address the above two issues. An illustration of our MMA is shown in Fig \ref{structure}. MMA designs different attention mechanisms for feature representation of point clouds of different scales during density changes. Specifically, in the Adjacency Attention Aggregation (AAA) module, we calculate the adjacency attention in the neighborhood through the adjacency relationship, so as to distinguish attributes within the cluster. This process makes features have a shape tendency by delineating the existence areas of each attribute. We then use parallel MLP to extract common point cloud features and aggregate the two. We save the aggregated features and continuously extract features with adjacent information until the density reaches a minimum value. Next, we perform density recovery. Similarly, we first perform ordinary feature extraction through MLP and then cascade the Feature Disparity Calculation (FDC) module to perform disparity attention calculation on the aggregated features at the same density stage as saved before. This actually establishes a connection between different receptive fields, allowing the network to compare two features during density changes and receptive field shrinkage, and learn the network's focus migration in the scene. We continue to calculate disparity attention during the process of increasing density, and actually calculate the feature disparity at different scales from near to far. FDC solidifies the shape tendency brought by the AAA module and further refines the possible existence areas of objects due to the multi-scale information brought by multiple parallaxes. Shape tendency and existence area inference given by AAA and FDC minimize problems caused by a lack of shape guidance and label jitter.

To summarize, our contributions are as follows:

\begin{itemize}
\item We propose MMA, a novel plug-and-in point cloud feature representation module designed for weakly-supervised detection. MMA produces extracted point cloud features characterized by inherent shape tendencies, which effectively address the absence of guidance regarding object shape due to the lack of well-annotated labels. MMA can activate the fully-supervised network's ability to detect weak labels, and can also enhance a weakly-supervised network's performance.
\item We propose the AAA module and FDC module in order to extract feature representation that is suitable for expressing weak labels. The AAA module can incrementally learn attribute distinctions for each object independently by calculating the attention of each point in the adjacent locations, thus giving each object the inherent shape tendency. The shape tendency is further solidified by FDC, which learns the migration of focus from the shallow layer to the deep layer, and thus resisting interference caused by label jitter and lack of shape guidance.
\item Extensive experiments demonstrate that fully-supervised detectors with MMA can increase the accuracy of detecting the weak labels, from only 60$\%$ of the weakly-supervised detectors to 90$\%$. At the same time, MMA can convert the label jitter problem that originally plagues weak supervision into the source of data enhancement, and making the weakly-supervised detector with MMA achieves the state-of-the-art in the indoor point-level weakly-supervised detection.
\end{itemize}

\section{Related Work}

The problem of 3D object detection has received significant interest over the past decades. 3D Object detection is the task of obtaining the correct bounding box, angles, and label for objects in a depth map.  Based on the accuracy of labeling, it can be further divided into fully-supervised, weakly-supervised, or semi-supervised detection. Among them, fully-supervised methods are currently the mainstream approach. However, full supervision faces challenges due to the high cost of precise 3D annotations. Therefore, weakly-supervised 3D object detection has become a new direction of development. At the same time, given the complexity of data in 3D scenes, some methods use attention mechanisms from Transformer to build detectors that achieve higher accuracy. We restrict our literature review to weakly-supervised methods and defer the interested reader to excellent surveys on the topics such as Transformer.

\subsection{3D weakly-supervised object detection} 

With the success of PointNet/PointNet++ \cite{pointnet,pointnet2} in point cloud classification and segmentation, the use of point clouds for 3D object detection is becoming increasingly popular. While achieving better accuracy, the complex point cloud processing structure and expensive cost of 3D data annotation make 3D detection challenging to implement. To address this, weakly-supervised, semi-supervised, and unsupervised methods have been proposed to reduce annotation costs. However, unsupervised \cite{unsup1, unsup2} and semi-supervised methods \cite{semi1, semi2} do not fully utilize 3D data and usually introduce more complex network structures than supervised methods. Weakly-supervised methods detect objects using the weakened form of labels and their accuracy is close to that of fully-supervised methods.

To achieve this goal, the method Back to Reality (BR) has been found to be useful to 
detection of weak labels by Using only the center point of the object. To address the information loss resulting from labeling only the center point, BR leverages synthetic 3D shapes to transform the weak label into a fully annotated virtual scene, providing stronger supervision. It then uses the perfect virtual label to complement and enhance the real label. In more detail, BR initially assembles the 3D shape into a physically plausible virtual scene based on the rough scene layout derived from location-level annotations. Subsequently, a virtual-real domain adaptive approach is applied to refine the weak labels and oversee the detector training using the virtual scene.

In addition, a new method WeakM3D, has recently emerged, which mainly studies how to train a 3D target detection network by using the target point cloud corresponding to the 2D detection boxes as the supervision signal for weak Supervised learning. WeakM3D obtains supervised signals of relevant parameters from point clouds and designs a point-level loss balance based on point density, including geometric target point cloud alignment loss and ray tracing loss. The alignment network projects the point cloud onto the image coordinate system, extracting the point cloud within the 2D bounding box. Then, employ an unsupervised target clustering algorithm to identify the target point cloud, followed by the computation of the loss between the predicted bounding box and the target point cloud. This training strategy relies on alignment networks and is constrained by unsupervised point cloud clustering algorithms, making it fail to capture complex point features.
The above methods mainly include constructing pre-trained 3D box prediction networks or designing data augmentation for weak labels. These methods either utilize the characteristics of weak labels, such as point-level weak labels can serve as center positions to create virtual scenes, or digging the impact of point cloud Information on 3D box prediction. The former lacks universality, while the latter based on the large-parameter pre-trained network, is not as good as the former's performance. At present, there is a lack of a coordinated approach that absorbs the advantages of both.

\subsection{Transformer for point cloud feature extraction} 
With the success of Transformer for natural language processing (NLP), there have been several attempts to adapt the model to object detection. Recently, Transformer has been applied to computer vision. Some methods have tried to use attention in space and channels to extract more comprehensive image features and improve accuracy in 2D object detection. In 3D object detection, Transformer has been proposed to enhance point cloud classification and segmentation in a supervised manner. For example, the Point Cloud Transformer (PCT) utilizes four offset attention modules to extract point features. Point Transformer (PT) replaces the MultiLayer Perceptron (MLP) layer in PointNet++ with multi-head attention. These works show good performance in fully-supervised detection.

However, our comparative experiments reveal that introducing the existing Transformer-based point cloud representation for weakly-supervised detection yields only marginal improvements. This disparity in performance can be attributed to the distinct feature requirements for weakly-supervised detectors in comparison to their fully-supervised counterparts, largely shaped by variations in label formats.

Currently, Transformer-based feature representation networks mainly emphasize fully-supervised tasks and pay little attention to weakly-supervised tasks. The existing Transformer-based point cloud feature representation method uses a consistent attention mechanism to process point clouds of different scales and different attributes, making it hard to effectively solve the problems of label jitter and lack of shape traction. The field of weak supervision urgently needs a point cloud feature representation method that is specific for weak labels.

\begin{figure}[t]
\centering
\includegraphics[width=0.5\columnwidth]{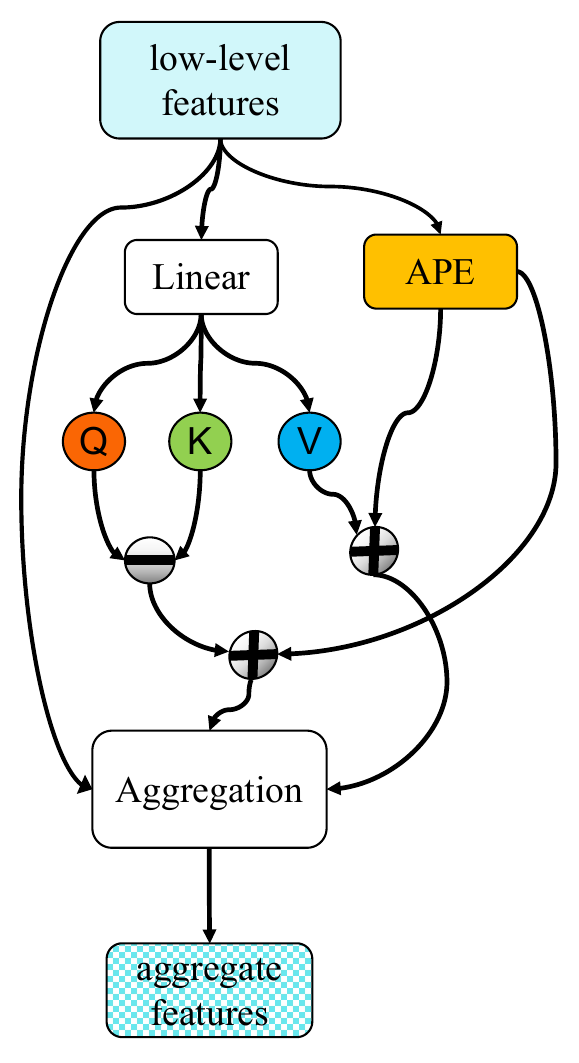}
\caption{The pipeline of Adjacency Attention Aggregation (AAA) module.}\label{AAA}
\end{figure}

\section{Methodology}
In this section, we will show the two important modules of our network: the Adjacency Attention Aggregation (AAA) module and the Feature Disparity Calculation Calculation (FDC) module. We will introduce their structure and how they work in the process of point cloud density reduction and recovery.

\subsection{Adjacency Attention branch}
The Adjacency Attention branch includes a position encoding layer named Adjacency Position Encoder (APE) and an adjacency attention layer. Among them, APE is a trainable module used to extract neighborhood relationships of local adjacent structures that help to understand local relationships.

\subsubsection{Adjacent Position Encoding (APE)}
In the Transformer, position coding is used to map the position information of word vectors. By inputting token location information, map it to the corresponding position encoding. This process is usually calculated by hard coding, such as binary vector marking or Periodic function.

For position coding, the 3D point cloud scene is similar to the environment of Natural language processing \cite{PA}. For point clouds and their local adjacent regions, their existence is similar to words and sentences. We establish adjacency mapping by encoding the positional relationships of neighboring points within a point cloud cluster. The APE module learns this mapping relationship to obtain the best representation of neighborhood relationships, which can be defined as:
\begin{equation}
\hfil \delta=\theta\left(p_i,p_j\right)
,   
\end{equation}

Where, $\delta$ represents the adjacency relationship representation that needs to be learned, and $p_i$ and $p_j$ represent the two adjacent nodes within a cluster that need to be mapped.

\subsubsection{Adjacency Attention Aggregation (AAA)}
Local relationships are an important component of point cloud relationships. Although point clouds contain rich depth and geometric information, they are irregular and lack sufficient local context compared to images. In the process of point cloud density decreasing, the point cloud decreases from complete and complex high density to streamlined low density. The adjacency relationship can continuously drive the filtering of point cloud local information in the process of density reduction, eliminate redundant information in the low-level feature of the high-density stage, and make the final streamlined point cloud have a certain shape tendency.

We use a self-attention mechanism to extract adjacency information in a local range, as shown in Fig \ref{AAA}. Specifically, we can define the following relationship:

\begin{equation}
\hfil y_i=\sum_{x_j \in X_{(i)}} \rho\left(\gamma\left(\varphi\left(x_i\right)-\psi\left(x_j\right)+\delta\right)\right) \cdot\left(\alpha\left(x_j\right)+\delta\right),   
\end{equation}

Where $X_{(i)}$ denotes the set of points in a local neighborhood of $x_i$, which can extract attention in the neighborhood around the point. $\varphi$, $\psi$, and $\alpha$ are feature transformations function. $\delta$ is the adjacency information learned from the APE module.  The output of the attention layer uses the subtraction between features transformed by $\varphi$ and $\psi$, which denotes the gap between two vectors, and the aggregating features transformed by $\alpha$. $\gamma$ is the mapping function and $\rho$ is a normalization function.

\subsection{Feature Disparity Attention branch}

In previous work, features are usually represented in the absolute space of point clouds. Point cloud features will continue to shrink the receptive field due to the progress of extraction work. An effective method is to expand the field of view through residual connections \cite{res}. However, only performing a simple residual connection operation on shallow features and deep features can only locally alleviate the problem of parallax. In MMA, we perform the extraction in the density restoration stage by establishing disparity attention for the same density stage before and after, as shown in Fig \ref{FDC}. FDC module focuses its attention on the feature disparity of the same density. By learning the difference in feature disparity, it can effectively learn the migration of network focus in the feature extraction process. Specifically, FDC can be expressed as:

\begin{figure}[b]
\centering
\includegraphics[width=0.55\columnwidth]{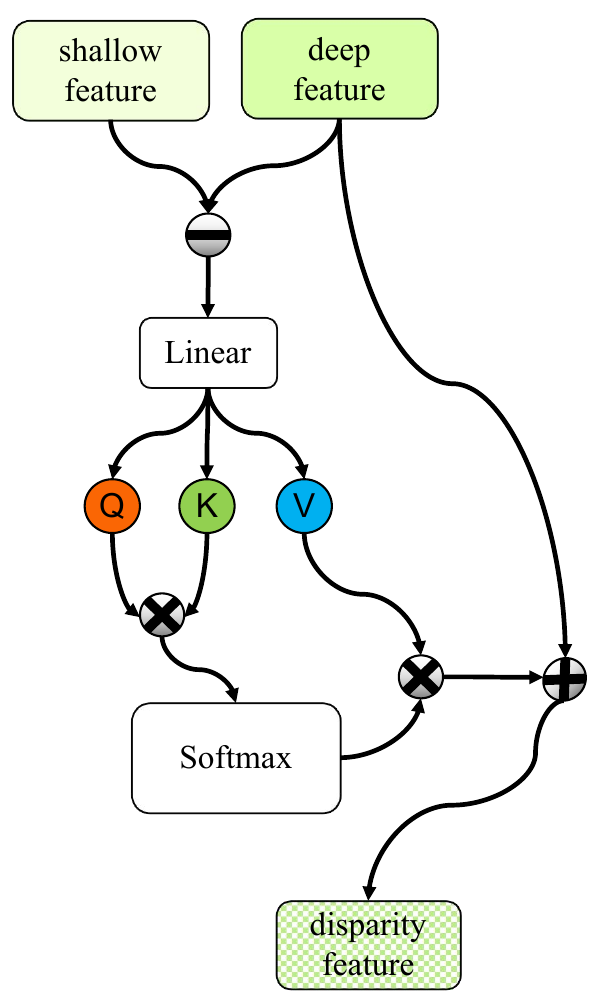}
\caption{The pipeline of Feature Disparity Calculation (FDC) module.}\label{FDC}
\end{figure}

\begin{equation}
y=(I-A) \cdot \beta\left(y_2-y_1\right),
\end{equation}

Among them, $y_2$ represents the connected shallow layer features, and $y_1$ represents the features extracted by the current MLP layer. Since $y_2$ and $y_1$ are at the same density level, they have the same number of point clouds. $\beta$ is element-wise subtraction, used to calculate the feature difference of different disparity feature maps. $I$ is an identity matrix comparable to the diagonal degree matrix D of the Laplacian matrix and $A$ is the attention matrix comparable to the adjacency matrix E.

\section{Experiment}\label{sec:experiment}
In order to verify the effect of MMA on weakly-supervised detection, we set up two parts of experiments: the experiment to activate the network's weakly-supervised detection ability and the experiment to improve the existing weakly-supervised detection ability. For the first part, we replace the point cloud feature extraction network of some commonly used detection networks, Group-Free3D \cite{GF3D} and VoteNet \cite{votenet}, with MMA for weakly-supervised detection. For the second part, we embed MMA into the current state-of-the-art indoor weakly-supervised detection network BR and observe its improvement in weakly-supervised detection networks. Besides, we set up rich ablation experiments to demonstrate the effectiveness of the structure.

\subsection{Experimental setup}

To ensure a fair comparison of the experimental results, we have adopted two current mainstream publicly available 3D datasets. One is  ScanNet \cite{scannet}, and another is Matterport3D \cite{matterport}. ScanNet is an RGB-D video dataset that includes 2.5 million views in 1201 training scenes and 312 validation scenes, annotated with 3D camera poses, surface reconstructions, and instance-level semantic segmentation. Matterport3D is a large-scale RGB-D dataset that contains 10,800 panoramic views from 194,400 RGB-D images of 90 building-scale scenes. 

We also performed extensive ablation experiments to demonstrate the necessity of the current structure by changing the internal structure of MMA. At the same time, we compared MMA with some other current point cloud feature extraction networks to prove the effectiveness of MMA. In addition, we also conducted an important experiment to compare the detection accuracy of the original network and the network embedded with MMA under different degrees of label jitter. Finally, due to the improvement of MMA, the subsequent detection network can save anti-jitter plug-ins and some detection heads, so the network structure can be simplified and network parameters can be reduced. Therefore, we tested the parameters when different networks obtained the best results after embedding MMA.

\subsection{Result and analysis}
In this section, we will show and analyze the experimental results. Our proposed MMA has improved the detector through all three aspects: accuracy, robustness, and model size. Besides, we performed ablation experiments to fully demonstrate the effectiveness of our structure.

\subsubsection{Accuracy}

\begin{table*}[htbp]
\centering
\caption{\textbf{The class-specific detection results with different detectors on ScanNet.} We denote the fully supervised baseline as FSB. We use $\--$MMA to represent the network with MMA and use the bold to represent the best result of all methods.}
\begin{adjustbox}{width=1.0\textwidth,height=0.07\textheight}
\begin{tabular}{|c|c|c|c|c|c|c|c|c|c|c|c|c|c|c|c|c|c|c|c|c|c|c|c|c|}
\hline
                             & settings                                                  & bath.         & bed           & bench         & bsf.          & bot.         & chair         & cup          & curt.         & desk          & door          & dres.         & keyb.         & lamp          & lapt.         & monit.        & n.s           & plant         & sofa          & stool         & table         & toils         & ward.         & mAP@0.25      \\ \hline
\multirow{8}{*}{ScanNet@0.25} & FSB                                                       & \textbf{86.2} & 87.5          & 16.3          & \textbf{49.6} & 0.6          & \textbf{92.5} & 0.0          & \textbf{70.9} & \textbf{78.5} & \textbf{53.5} & \textbf{56.0} & 6.4           & \textbf{68.2} & 11.5          & 81.5          & 88.5          & 15.2          & 88.2          & 45.6          & \textbf{65.0} & \textbf{99.7} & \textbf{31.2} & \textbf{54.2} \\ \cline{2-25} 
                                 &VNet     
                                        &5.0 &30.2 &2.6 &5.5 &0.0 &64.0 &0.0 &1.7 &26.7 &1.7 &0.1 &1.4 &0.0 &0.8 &24.7 &0.2 &3.0 &49.7 &0.5 &2.7 &0.7 &0.2 &10.1 \\ \cline{2-25}              
                                &VNet-MMA     
                                        &39.7 &87.9 &4.6 &14.3 &0.0 &68.5 &0.0 &3.0 &20.5 &1.9 &8.1 &2.4 &11.7 &0.8 &33.0 &33.0 &1.2 &66.5 &9.4 &12.4 &90.3 &2.8 &23.3 \\ \cline{2-25}                         
                             & GF3D                                                       & 45.0          & 75.7          & 4.3           & 17.2          & 0.0          & 81.4          & 0.0          & 3.5           & 34.0          & 4.7           & 3.2           & 2.1           & 46.6          & 3.3           & 45.8          & 52.8          & 8.3           & 71.0          & 15.7          & 18.1          & 90.8          & 0.7           & 29.7          \\ \cline{2-25} 
                             & GF3D-MMA                                                    & 84.9          & 84.3          & 3.9           & 26.8          & 0.0          & 82.5          & 0.0          & 13.0          & 59.8          & 10.2          & 52.4          & 15.5          & 53.7          & 6.5           & 70.3          & 90.9          & 5.4           & 84.4          & 33.3          & 34.8          & 99.6          & 5.5           &  41.7    \\ \cline{2-25} 
                             & BRM                                                       & 85.3          & \textbf{90.9} & 8.8           & 34.3          & 1.9          & 80.0          & \textbf{7.7} & 24.7          & 58.0          & 20.8          & 45.4          & \textbf{31.3} & 64.4          & 25.8          & 67.5          & 76.7          & 27.3          & \textbf{91.4} & 43.3          & 46.7          & 94.8          & 8.3           & 47.1          \\ \cline{2-25} 
                             & BRM-MMA                                                    & 69.8          & 87.8          & 20.3          & 39.0          & \textbf{4.7} & 83.2          & 5.0          & 37.2          & 66.9          & 28.8          & 31.9          & 12.7          & 58.5          & 29.2          & \textbf{98.6} & \textbf{94.0} & \textbf{27.6} & 84.1          & 47.3          & 47.4          & 98.8          & 3.1           &  47.5    \\ \cline{2-25} 
                             & \begin{tabular}[c]{@{}c@{}}BRM-MMA\\ (jitter)\end{tabular} & 81.2          & 83.1          & \textbf{43.8} & 43.3          & 0.6          & 85.0          & 0.2          & 44.8          & 62.9          & 27.3          & 43.3          & 10.6          & 56.3          & \textbf{32.9} & 54.5          & 83.7          & 23.6          & 86.9          & \textbf{51.0} & 47.4          & 93.4          & 2.0           &  48.1    \\ \hline
\end{tabular}%
\end{adjustbox}
\label{scannet025}
\end{table*}

\begin{table*}[htbp]
\centering
\caption{\textbf{The class-specific detection results with different detectors on Matterport.}}
\begin{adjustbox}{width=1.0\textwidth,height=0.08\textheight}
\begin{tabular}{|c|c|c|c|c|c|c|c|c|c|c|c|c|c|c|c|}
\hline
\textbf{}                    & settings                                                  & bath.          & bed           & bench         & chair         & curt.         & desk          & door          & dres.         & n.s.          & sofa          & stool         & table         & tois          & mAP@0.25      \\ \hline
\multirow{8}{*}{Matterport3D@0.25} & FSB                                                       & 100.0          & 90.3          & 29.7          & 71.1          & \textbf{12.9} & 6.3           & \textbf{13.1} & 1.6           & 62.4          & 65.1          & 18.2 & \textbf{47.1} & 75.9          & \textbf{45.7} \\ \cline{2-16} 
                             & VNet & 53.3 & 1.4 & 2.1 & 53.9 & 1.7 & 10.0 & 3.4 & 1.4 & 1.5 & 34.0 & 16.9 & 7.8 & 53.7 & 18.5 \\ \cline{2-16}
                             & VNet-MMA & 59.8 & 47.2 & 0.3 & 45.1 & 0.2 & 2.2 & 8.3 & 8.2 & 40.5 & 45.2 & 6.2 & 21.2 & 67.4 & 27.1 \\ \cline{2-16}              
                             & GF3D & 75.0 & 55.7 & 16.8 & 17.8 & 0.0 & 0.0 & 1.5 & 0.0 & 1.2 & 36.4 & 0.3 & 1.4 & 82.0 & 22.2 \\ \cline{2-16}
                             & GF3D-MMA & 100.0 & 73.9 & 30.0 & 58.3 & 0.8 & 10.0 & 8.6 & 0.4 & 59.0 & 57.1 & \textbf{33.5} & 23.8 & 78.1 & 41.0 \\ \cline{2-16}                             
                             & BRM                                                       & 95.0           & 72.0          & \textbf{37.0} & 58.3          & 2.1           & \textbf{20.1} & 7.2           & 3.0           & \textbf{63.1} & \textbf{68.3} & 12.7          & 27.7          & 76.1          & 41.7          \\ \cline{2-16} 
                             & BRM-MMA                                                    & 100.0          & 87.0          & 36.0          & 64.9          & 6.2           & 3.2           & 8.7           & \textbf{16.5} & 53.5          & 59.7          & 9.8           & 21.1          & 70.2          &  41.9    \\ \cline{2-16} 
                             & \begin{tabular}[c]{@{}c@{}}BRM-MMA\\ (jitter)\end{tabular} & \textbf{100.0} & \textbf{90.5} & 26.3          & \textbf{73.1} & 7.3           & 7.3           & 8.9           & 1.8           & 50.2          & 67.7          & 15.4          & 27.7          & \textbf{84.1} & 43.1    \\ \hline

\end{tabular}%
\end{adjustbox}\vspace{5pt}
\label{matterport025}
\end{table*}

We replace the point cloud feature extractor in the current mainstream 3D object detection networks Group-Free3D (GF3D) and VoteNet (VNet) with MMA and observe the network's ability to detect weak labels. At the same time, we also embed MMA into the current state-of-the-art indoor weakly-supervised detection framework BR to observe the improvement on weakly-supervised-targeted detectors, as shown in Table \ref{scannet025} and Table \ref{matterport025}.

We first use those mainstream fully-supervised detectors to perform weak supervision detection tasks directly. The results of GF3D and VNet on ScanNet and Matterport show that non-special designed detectors are poor at handling weak labels, only achieving 29.7$\%$ and 10.1$\%$ accuracy on ScanNet, 22.2$\%$ and 18.5$\%$ on Matterport. GF3D only achieves about 50$\%$ to 60$\%$ of BRM accuracy, and even lower for VNet which is about 20$\%$ to 40$\%$. This phenomenon indicates the performance of some existing fully-supervised detectors is insufficient to support their weakly-supervised detection. Then, we embed the MMA into fully supervised detectors and find a huge improvement from the baseline. GF3D achieves 41.7$\%$ and 41.0$\%$ on ScanNet and Matterport, while VNet achieves 23.3$\%$ and 27.1$\%$. GF3D-MMA even maintains a similar performance as BR.

On the other side, we chose the BRM model, which achieves the highest accuracy among BR models, as our embedding object. The embedded BRM is called BRM-MMA. BRM-MMA achieves higher mAP on the ScanNet and Matterport3D datasets with 48.1 $\%$ (54.2$\%$ for full supervision) and 43.1$\%$ (45.7 for full supervision), which is higher than baseline for 1$\%$ and 1.4$\%$. These results mean we could achieve 88.7$\%$ and 94$\%$ of full supervision performance while only requiring about 5$\%$ annotation time.

\subsubsection{Robustness and model Size}

\begin{figure}[t]
    \centering
    \includegraphics[width=0.8\columnwidth]{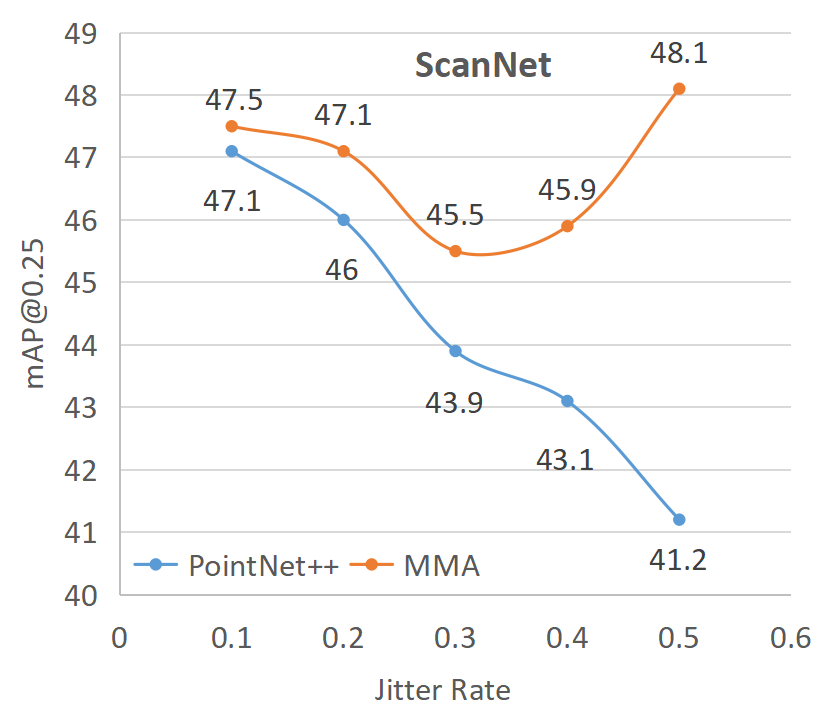}
    % \input{FCN8.tex}
    % \subimport{latexpic/HRNet.tex}
    \caption{Variation trend of accuracy with jitter on ScanNet.} %caption title
    \label{jitter1}
\end{figure}

\begin{figure}[t]
    \centering
    \includegraphics[width=0.75\columnwidth]{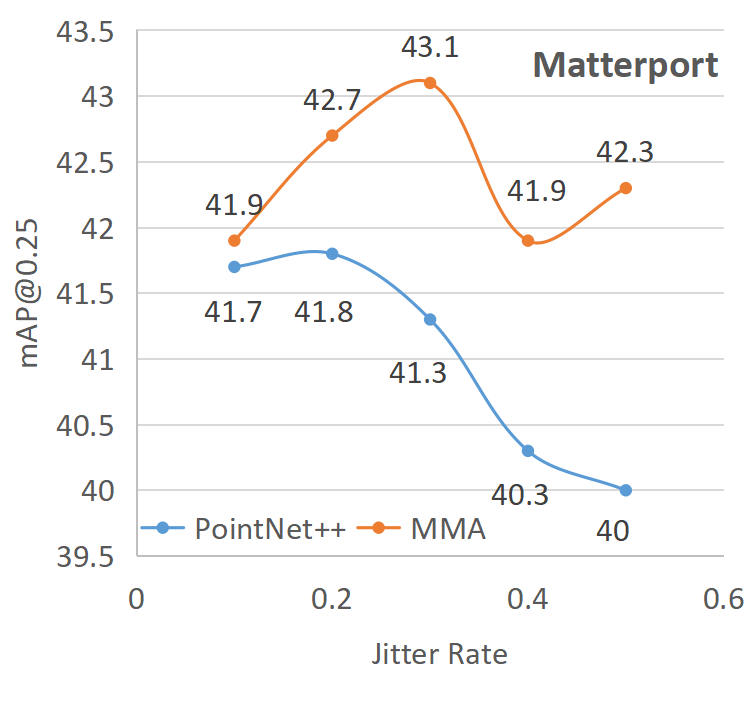}
    % \input{FCN8.tex}
    % \subimport{latexpic/HRNet.tex}
    \caption{Variation trend of accuracy with jitter on matterport.} %caption title
    \label{jitter2}
\end{figure}

Due to the robust feature brought by shape tendency, MMA introduces a magical anti-jitter effect, so that the label jitter that originally affected the accuracy becomes the source of label enhancement, making the detection accuracy reach even higher. The accuracy-jitter curves are shown in Fig \ref{jitter1} and Fig \ref{jitter2}.

We can notice that the curves on Matterport are different from ScanNet. The variation curve on ScanNet is a Concave function, while Matterport is more similar to a Convex function. This may be because ScanNet is dominated by small objects. The small scale of objects brings little absolute difference between the error label and the true label, which is not significant in the entire scene. This allows MMA to utilize high-coefficient error information on ScanNet. While on Matterport, there are mainly large objects. The high-coefficient error of these large objects may confuse the entire scene, making it difficult to utilize. Therefore, Matterport performs better under a relatively moderate 30$\%$ jitter.

\begin{figure}[t]
    \centering
    \includegraphics[width=0.85\columnwidth]{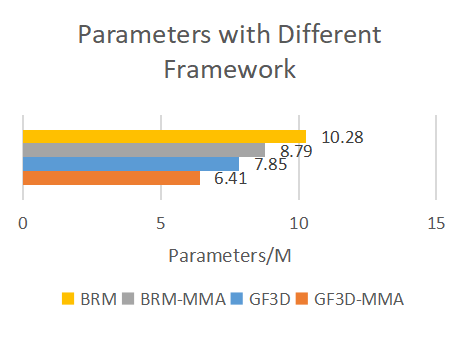}
    % \input{FCN8.tex}
    % \subimport{latexpic/HRNet.tex}
    \caption{The number of parameters for different models.} %caption title
    \label{para}
\end{figure}

Besides, due to the robustness brought by MMA, we trim unnecessary center refinement modules in BRM, which were originally used to anti-jitter. For the reduction of relevant refining modules, the overall parameter quantity of BRM-MMA has also significantly decreased compared to BRM, from 10.28M to 8.79M. At the same time, since point cloud features refine the information, detection networks using multi-head attention such as GroupFree-3D can achieve the highest accuracy with fewer heads. The parameter amount of GF3D-MMA is reduced to 6.41M compared with 7.85M of GF3D, as shown in Fig \ref{para}.

\subsection{Ablation experiment}
We conduct ablation experiments to explore the effectiveness of each module. We first explore the impact of the number of modules. In our model structure, we use 3-5 times density reduction and 2-4 times density recovery with a corresponding number of AAA and FDC modules to detect the impact of the number of modules on the final results. Experiments show that the 4+3 model is the most effective. The results are shown in the Table \ref{numberblock}.

\begin{table}[htbp]
\centering
    \caption{Ablation results of different module quantities}{\adjustbox{width=.6\linewidth,height=0.04\textheight}{
    \begin{tabular}{l|l|l|l}
      \hline
           & 3*AAA & 4*AAA          & 5*AAA \\ \hline
      2*FDC & 44.9 & 47.7          & 44.5 \\ \hline
      3*FDC & 46.1 & \textbf{48.1} & 43.8 \\ \hline
      4*FDC & 46.0 & 45.2          & 43.2 \\ \hline
    \end{tabular}
  }}
  \label{numberblock}
\end{table}

\begin{figure}[htbp]
    \centering
    \includegraphics[width=1\columnwidth]{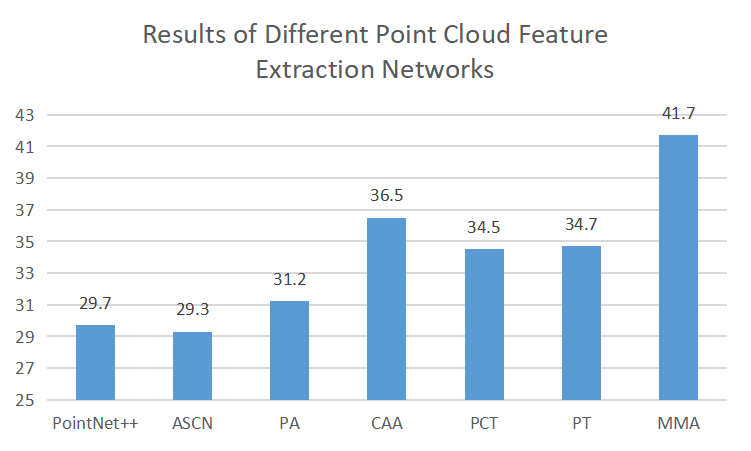}
    % \input{FCN8.tex}
    % \subimport{latexpic/HRNet.tex}
    \caption{The results of replacing GF3D with various point cloud extraction networks for weakly-supervised object detection on ScanNet.} %caption title
    \label{dbackbone}  
\end{figure}

We compare our structure with many other point cloud extraction networks, some of which are also based on attention mechanisms. The result is shown in Fig \ref{dbackbone}. All six networks are PointNet++, Attention ShapeContextNet (ASCN) \cite{ASCN}, Point Attention (PA) \cite{PA}, Channel wise Affinity Attention (CAA) \cite{CAA}, Point Cloud Transformer (PCT) \cite{PCT}, and Point Transformer (PT) \cite{PT}.

\section{Visualization}
\begin{figure*}[htbp]
    \centering
    \includegraphics[width=1\textwidth]{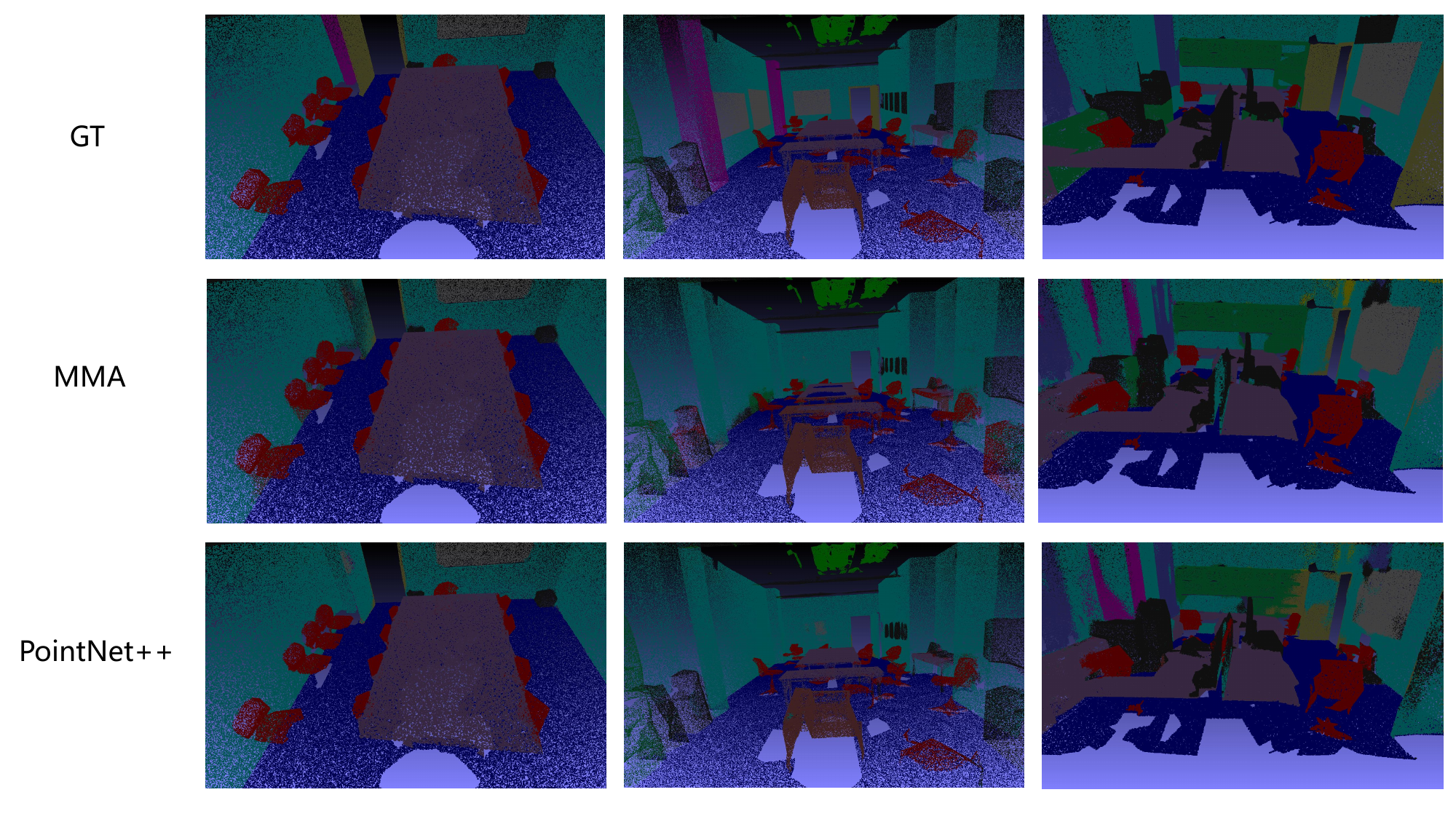}
    \caption{\textbf{Semantic segmentation results on S3DIS}} %caption title
    \label{vis1}
\end{figure*}

Visualization results of semantic segmentation using MMA are in Fig \ref{vis1}. We embed MMA into pointnet2/SSG and use S3DIS as our dataset. We found that the MMA overall Acc and class avg IoU have increased from 83.0$\%$ and 53.5$\%$ to 84.8$\%$ and 65.3$\%$ compared to PointNet++, respectively.

\section{Conclusion}\label{sec:conclusion}
We propose a plug-and-in point cloud feature extraction network MMA that adopts different attention strategies according to different density change trends and establishes feature learning for multiple disparities. We use AAA and FDC modules to learn intra-cluster adjacency relationships and the focus transfer of the network when the receptive field is reduced so that the learned features have shape tendencies and shape stability. In the case of indoor point labels, MMA can activate the weak supervision capabilities of mainstream detectors only through changes in point cloud feature extraction. When MMA is embedded into a weakly-supervised detection network, it can also enhance the existing detection capabilities. At the same time, MMA can help reduce the structure of the network to a certain extent, reducing the amount of network parameters. For weakly supervised detection networks, MMA can help the network resist the interference of label jitter, and to a certain extent, even convert label jitter into a source of data enhancement.

\end{document}